\documentclass{article}
\usepackage[utf8]{inputenc}
\usepackage{algorithm2e}
\usepackage{algorithmic}
\usepackage{xcolor}
\usepackage{listings}
\usepackage{url}
\usepackage{hyperref}
\usepackage{pgfplots}
\usepackage{float}
\usepackage{booktabs}
\usepackage{adjustbox}
\usepackage{multirow}
\usepackage{booktabs}
\usepackage{rotating}
\usepackage{amsmath,amssymb,amsfonts}
\usepackage{cite}
\usepackage{array}
\usepackage[english]{babel}
\newenvironment{keywords}{%
    \par\medskip\noindent%
    \textbf{Keywords:}\ \ignorespaces%
}{%
    \par\medskip%
}
\title{An experimental approach on Few Shot Class Incremental Learning}

\author{Marinela Adam}

\date{10 June 2024}
\pgfplotsset{compat=1.18}
\begin{document}

\maketitle

\begin{abstract}
Few-Shot Class-Incremental Learning (FSCIL) represents a cutting-edge paradigm within the broader scope of machine learning, designed to empower models with the ability to assimilate new classes of data with limited examples while safeguarding existing knowledge. The paper will present different solutions which contain extensive experiments across large-scale datasets, domain shifts, and network architectures to evaluate and compare the selected methods. We highlight their advantages and then present an experimental approach with the purpose of improving the most promising one by replacing the visual-language (V-L) model (CLIP) with another V-L model (CLOOB) that seem to outperform it on zero-shot learning tasks. The aim of this report is to present an experimental method for FSCIL that would improve its performance. We also plan to offer an overview followed by an analysis of the recent advancements in FSCIL domain, focusing on various strategies to mitigate catastrophic forgetting and improve the adaptability of models to evolving tasks and datasets. 
\end{abstract}


\section{Introduction}
Few-Shot Class-Incremental Learning (FSCIL), as discussed in \cite{FewShotClassIncremental2020-xiaoyutao}, represents a pioneering approach in machine learning, enabling models to learn new classes from a limited number of examples while retaining previously acquired knowledge. Inspired by the human ability to progressively accumulate knowledge, FSCIL tackles the challenge of adapting machine learning models to continually evolving datasets. This advanced method allows models to generalize and learn effectively from sparse instances, emulating the cognitive processes observed in human learning.
\subsection{Motivation}
In the rapidly evolving field of machine learning, one of the most challenging tasks is to develop models that can learn from a small number of examples \cite{sarker2021deep}, paradigm known as Few-Shot Learning (FSL). FSL, as presented in \cite{snell2017prototypical}, has shown promising results in various domains, including image recognition, natural language processing, and reinforcement learning. However, these models often assume that all classes are available during training, which is not always the case in real-world scenarios.

In many practical applications, new classes can emerge over time, and models need to adapt to these changes. This scenario is referred to as Class-Incremental Learning \cite{mittal2021essentials} (CIL). While CIL has been extensively studied \cite{onchics2021double}, \cite{hogea2024fetril++}, \cite{slim2022dataset}, most existing approaches suffer from a phenomenon known as “catastrophic forgetting” \cite{kirkpatrick2017overcoming}, where the model’s performance on the original classes deteriorates when learning new ones. This aligns with recent advancements in incremental learning techniques, such as controlling dark knowledge values for optimal knowledge distillation, as presented in \cite{onchis2024optimal}, which aim to mitigate catastrophic forgetting through non-heuristic mechanisms.

The intersection of FSL and CIL, known as Few-Shot Class-Incremental Learning (FSCIL), is a relatively unexplored area of research. FSCIL aims to develop models that can learn new classes from a few examples without forgetting the previously learned ones. 

This capability is crucial for many real-world applications, such as medical diagnosis, where new diseases (classes) can emerge, and only a few examples might be available initially.
\subsection{Context of the research}
The contextual landscape of Few-Shot Class-Incremental Learning (FSCIL) revolves, as presented in \cite{FewShotClassIncremental2020-xiaoyutao}, around fundamental research inquiries. Central questions in this domain include investigating how models can assimilate new classes from a limited number of examples without erasing previously acquired knowledge, a challenge often referred to as “catastrophic forgetting”. This involves striking a balance between stability and plasticity, which is the equilibrium between retaining existing knowledge while acquiring novel information.

FSCIL also explores how models can harness prior knowledge and transfer learning to enhance performance when faced with new classes. This mirrors the learning processes observed in humans and is a crucial aspect of FSCIL. Understanding how these models adapt to shifts in data distribution and evolving class complexities over time is another avenue of inquiry in this field. 

Real-world challenges that stand to benefit from FSCIL methodologies are diverse, as presented in the survey \cite{tian2024survey}. In the realm of medical diagnosis \cite{sun2023few}, for instance, FSCIL can facilitate model adaptation to new diseases (classes), enhancing their efficacy in diagnosis. In user profiling scenarios, models leverage FSCIL to assimilate user feedback and preferences, enabling the provision of personalized recommendations. FSCIL proves advantageous in fraud detection by empowering models to discern new patterns of fraudulent behavior and prevent financial losses. Additionally, in fields like weather prediction and news recommendation, FSCIL contributes by enabling models to assimilate new classes, enhancing their accuracy, reliability, and relevance.

The historical roots of incremental learning extend back to the nascent days of artificial neural networks and machine learning. Pioneering works in this field, such as Grossberg (1987) \cite{GROSSBERG198723}, and Schlimmer and Granger Jr (1986) \cite{schlimmer1986incremental}, have significantly shaped the trajectory of research in incremental learning. These foundational contributions laid the groundwork for subsequent developments in the field, showcasing the enduring relevance and evolution of incremental learning methodologies. The emergence of FSCIL as a distinct area of research represents a significant evolution in this trajectory, addressing the unique challenges posed by few-shot, class-incremental scenarios.
\subsection{Main Research Questions}
The main questions that started this work were obtained during research of different approaches on FSCIL domain. We found a promising approach with very good results (LP-DiF \cite{huang2024learning}) that seemed to outperform the SOTA methods such as LIMIT \cite{FS_CIL_SamplingMultiPhaseTasks2022-daweizhou} and ERDIL \cite{FS_CIL_RelationKnowledgeDistillation2021-songlindong}. We delved into it and we found it has three main components, one of them being a vision-language model (CLIP \cite{pmlr-v139-radford21a}) which was used for zero-shot learning (ZSL) capabilities. During the process of analyzing it we found a new approach (CLOOB \cite{fürst2022cloob}) that seemed to outperform it and so we continued the research in this direction.
The main questions that started this paper are:
\begin{itemize}
    \item Which are the best methods to mitigate catastrophic forgetting?
    \item Is it possible to integrate the new architecture of CLOOB in LP-DiF approach?
    \item Will the replacement of CLIP with CLOOB improve LP-DiF accuracy?
\end{itemize}
\subsection{Structure of the report}
In the rest of the sections of the paper will be presented the following aspects.

Section II will present the Methodology of research with accent on the workflow of the approach and the theoretical analysis.

Section III will delve in the Proposed approach, presenting the datasets used, the starting point of this approach (LP-DiF) and the elements that will be used.

Section IV will present the preliminary results consisting of initial results of LP-DiF on FSCIL task, CLIP and CLOOB on Zero-Shot Learning task.

Section V will present the main results obtained with our proposed approach, will compare them with similar solutions and then with other state of the art approaches for FSCIL.

Section VI will start with the conclusion of our work and what was presented so far, then some limitations and open problems will be presented and completed with future work directions that should mitigate some of the open problems.

\section{Methodology of research}
\subsection{Theoretical analysis}
We started our research with the theoretical analysis that consists of studying different incremental learning methods. We conducted a literature review where we analyzed several different approaches on FSCIL (TOPIC \cite{FewShotClassIncremental2020-xiaoyutao}, LIMIT\cite{FS_CIL_SamplingMultiPhaseTasks2022-daweizhou}, ERDIL\cite{FS_CIL_RelationKnowledgeDistillation2021-songlindong}, LP-DiF\cite{huang2024learning}, SV-T\cite{qiu2023semantic}, found their strong and weak points, noted their core elements and found the challenges they faced (catastrophic forgetting). We decided to study in depth some FSCIL approaches and try improve one of them (LP-DiF). One of the components that were used in LP-DiF was CLIP, so we moved the study in this direction to see if we can find a better approach that would also improve the FSCIL framework. 

\subsection{Approach workflow}
We started with a workflow plan that consists of the following steps:
\begin{enumerate}
    \item train and evaluate LP-DiF approach on three datasets (CIFAR100, miniImageNet and CUB200)
    \item train and evaluate CLIP model (used provided model checkpoints)
    \item train and evaluate CLOOB model (used provided model checkpoints)
    \item try to integrate CLOOB in LP-DiF
    \item replace CLIP with CLOOB in LP-DiF and retrain
    \item compared the original approach with ours and with other SOTA ones
\end{enumerate}
With this plan we wish to see if we can obtain better results on FSCIL by leveraging the progress in ZSL area and how does the model change affect the overall architecture.
\section{Proposed Approach}
 Out proposed approach plans to introduce an improvement to the Learning Prompt with Distribution-based Feature Replay (LP-DiF) framework to boost its performance in few-shot class-incremental learning situations. The LP-DiF framework usually uses the CLIP as vision-language model, which has been successful in reducing catastrophic forgetting by using learnable prompts and pseudo-feature replay. However, CLIP has some limitations, especially with feature saturation and the explaining away problem. To tackle these issues, we propose replacing CLIP with CLOOB, a model designed to be more robust and perform better using advanced contrastive learning techniques and having a better representation of features. By integrating CLOOB into the LP-DiF framework, we aim to assess its effectiveness compared to CLIP, focusing on its ability to gradually learn and retain knowledge.
\subsection{The LP-DiF Framework}

The LP-DiF framework is an innovative and sophisticated approach designed for Few-Shot Class-Incremental Learning (FSCIL). It integrates the strengths of the CLIP model along other innovative techniques to address the core challenges of incremental learning, namely catastrophic forgetting and limited training data. The main components of the LP-DiF framework are the Variational Autoencoder (VAE), prompt tuning, and feature replay. Each of these components plays a critical role in enhancing the framework's performance and efficiency and so we'll further investigate them.

\paragraph{Variational Autoencoder (VAE)}

The VAE in the LP-DiF framework is used for generating pseudo-features of old classes to preserve their feature distributions during incremental learning. The VAE consists of an encoder and a decoder:

\begin{itemize}
    \item \textbf{Encoder:} The encoder processes image features extracted by the vision-language model (CLIP), transforming them into a latent code \( z \). This latent code is modeled to follow a Gaussian distribution \( \mathcal{N}(0, I) \), ensuring a smooth and continuous latent space that facilitates reliable feature generation.
    \item \textbf{Decoder:} The decoder uses the latent code \( z \) to generate synthetic features. These synthetic features are combined with learnable prompts to produce pseudo-features that mimic the real features of the old classes.
\end{itemize}

The VAE is trained using a hybrid loss composed of two loss functions:
\begin{itemize}
    \item \textbf{Reconstruction Loss} (\( \mathcal{L}_r \)): This loss measures the difference between the original features and the reconstructed features, ensuring that the generated features accurately represent the original data.
    \item \textbf{Kullback-Leibler Divergence} (\( \mathcal{L}_{KL} \)): This regularizes the latent space to conform to the Gaussian prior, promoting smooth and diverse feature generation.
\end{itemize}

The total VAE loss is a weighted sum of these two losses: 
\[ \mathcal{L}_{VAE} = \mathcal{L}_{KL} + \lambda_r \mathcal{L}_r \]
where \( \lambda_r \) is a balancing coefficient.

\paragraph{Prompt Tuning}

Prompt tuning in LP-DiF adapts the vision-language model (CLIP) to new tasks while retaining knowledge from previous tasks. This involves:

\begin{enumerate}
    \item \textbf{Initialization:} At the beginning of each incremental learning session, the prompt vectors are initialized based on the weights learned from the previous sessions.
    \item \textbf{Training:} The prompts are trained using both the current session's training data and pseudo-features from the old classes generated by the VAE. This dual training approach ensures that the model learns new information without forgetting the old.
\end{enumerate}

By adjusting the prompts, the framework can effectively integrate new class information while maintaining the performance on previously learned classes.

\paragraph{Feature Replay}

Feature replay in LP-DiF addresses the challenge of catastrophic forgetting by replaying pseudo-features instead of raw images. This process involves:

\begin{itemize}
    \item \textbf{Feature-Level Distributions:} Instead of storing raw images, LP-DiF stores feature-level distributions, including the mean and variance vectors for each class. This significantly reduces storage requirements.
    \item \textbf{Pseudo-Feature Generation:} Pseudo-features are generated from the stored distributions using the VAE. These pseudo-features are used during training to represent old classes, allowing the model to retain knowledge of previous classes while learning new ones.
\end{itemize}

This method ensures that the model does not forget previously learned information, maintaining high accuracy across all classes over successive learning sessions.

\subsection{CLIP}
CLIP (Contrastive Language-Image Pretraining) \cite{pmlr-v139-radford21a} is a vision-language model that learns to associate images and text by jointly training on a large dataset of images and their descriptions. In LP-DiF, CLIP's image encoder extracts robust feature representations, while the text encoder provides corresponding textual descriptions. This combination allows for effective zero-shot and few-shot learning, significantly enhancing FSCIL performance.

\subsection{CLOOB}
CLOOB (Contrastive Leave One Out Boost) \cite{fürst2022cloob} builds upon CLIP by incorporating InfoLOOB, a technique that enhances contrastive learning through improved negative sampling. CLOOB's architecture also utilizes vision-language models, similar to CLIP, but aims to improve robustness and performance in zero-shot and few-shot scenarios. In our experiments, we will replace CLIP with CLOOB in the LP-DiF framework to evaluate its impact on FSCIL, having in mind that in the tests presented by the authors, it managed to outperform CLIP in most cases.

\subsection{Datasets}
Now we will shortly present the main datasets that are used for our work and on which were used to compare different approaches.
\subsubsection{YFCC Dataset}
The Yahoo Flickr Creative Commons dataset is utilized for training both CLIP and CLOOB models. It contains millions of images and their corresponding metadata, providing a vast and diverse dataset crucial for pretraining robust vision-language models. The dataset includes various categories of images, allowing the models to learn comprehensive visual and textual representations.

\subsubsection{LP-DiF Benchmark Datasets}
For evaluating the LP-DiF framework with CLIP and CLOOB, we will consider following benchmark datasets:

\paragraph{CIFAR-100}
CIFAR-100 consists of 60,000 32x32 color images in 100 classes, with 600 images per class. There are 50,000 training images and 10,000 test images. Each class contains 500 training images and 100 testing images, providing a challenging benchmark for evaluating incremental learning methods.

\paragraph{mini-ImageNet}
mini-ImageNet is a subset of the ImageNet dataset, designed for few-shot learning research. It contains 100 classes with 600 images each, split into 64 training, 16 validation, and 20 test classes. The images are resized to 84x84 pixels, making it a suitable benchmark for few-shot class-incremental learning.

\paragraph{CUB-200}
The Caltech-UCSD Birds-200-2011 (CUB-200) dataset contains 11,788 images of 200 bird species. Each species is represented by approximately 60 images, annotated with bounding boxes, part locations, and attributes. This fine-grained dataset is used to evaluate the framework's ability to handle detailed and specific classes.

\subsubsection{Additional datasets}
\paragraph{SUN-397}
The SUN-397 dataset includes 108,754 images across 397 scene categories. This dataset is used to assess the performance of models in recognizing a wide variety of scenes, providing a comprehensive evaluation for the incremental learning framework.

\paragraph{CUB-200*}
The modified CUB-200* dataset is proposed in the LP-DiF framework to introduce more challenging conditions for incremental learning. It includes additional augmentations and variations, making it a stringent benchmark for evaluating the robustness and adaptability of the learning models.

\subsection{Our Approach}
We will use two versions of CLIP and CLOOB (those we have available checkpoints for), both trained on the YFCC dataset with RN50 and RN50x4 backbones, and adapt the LP-DiF framework to use both model's architectures. Specifically, CLIP and CLOOB will be integrated into the few-shot class-incremental learning (FSCIL) setup by replacing the vision-language model in the LP-DiF framework with these variants. This adaptation will allow us to evaluate the effectiveness of CLOOB compared to CLIP in maintaining and acquiring knowledge incrementally.

The paper \cite{huang2024learning} employed the integration of CLIP with ViT-B/16 architecture for experimental purposes. However, given the availability of pretrained models from the CLOOB approach, specifically, those utilizing CLIP and CLOOB trained on the YFCC dataset with ResNet-50 (RN50) and its 4x scaled version (RN50X4) as backbone architectures, we will conduct our experiments utilizing these pre-trained models. This decision is motivated by computational constraints, as retraining models with alternative backbone architectures is not feasible within our current computational resources.

The CLIP model is known for its impressive zero-shot transfer learning capabilities, leveraging a rich representation learned from the InfoNCE objective and natural language supervision. However, it suffers from the explaining away problem, where it focuses on one or few features while neglecting other relevant features. CLOOB addresses this issue by using modern Hopfield networks to enhance the covariance structure in the embeddings, combined with the InfoLOOB objective to mitigate the saturation effect observed with InfoNCE.
\section{Preliminary results}
In this section we will present the preliminary results that served as the starting point for our approach as well as the comparison element to quantify the effectiveness of our approach.
\subsection{LP-DiF}
Initial results with LP-DiF prove to be very promising, obtaining SOTA performance on several benchmark datasets like CIFAR100, miniImageNet and CUB-200. 

As a first point of the results, it shows that learning prompts (LP) outperform linear classifiers (LC) across the three benchmarks and incorporating old-class distribution (OCD) improves performance but still LP + OCD obtains best performance.

It shows on CUB200 dataset that it achieves better balance between base and new classes by outperforming competitors on both new and old classes.

LP-DiF outperforms other models like BiDistFSCIL on CUB200, regardless of shot numbers. It uses both real features and synthesized features for Gaussian distribution estimation to obtain better performance and proves in a different test that uses KL divergence that synthesized features enriches class-relevant information.

One of the best achievements of the model is obtaining an average accuracy on miniImageNet of 93.76\% surpassing CLIP+BDF by 9.13\% and getting close to the theoretical upper bound that is 94.81\%.

\begin{figure}[htbp]
    \centering
    \includegraphics[width=\columnwidth]{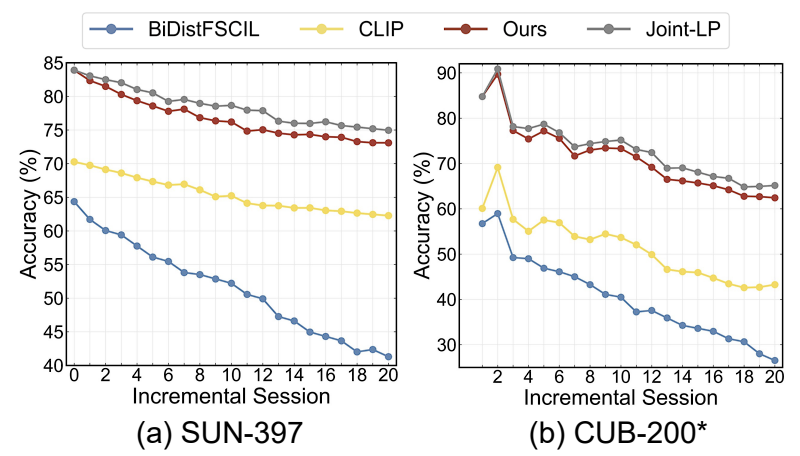}
    \caption{Accuracy curves LP-DiF and comparison with counterparts on SUN-397 and CUB200* datasets}
    \label{Figure1}
\end{figure}

Other results on different datasets (as average across all sessions) are:
\begin{itemize}
    \item CUB200: Using ViT-B/16 backbone obtains 74\% accuracy (+4.51\% more than second best accuracy)
    \item CIFAR100: Using ViT-B/16 backbone obtains 75.12\% accuracy (+4.26\% more than second best accuracy)
    \item best and comparable with theoretical upper bound on SUN-397 and CUB-200* (Fig. \ref{Figure1})
    \item less storage space required compared to methods like iCarRL
\end{itemize}
Some flaws that we identified are:
\begin{itemize}
    \item model is parameter sensitive, high values negatively influencing the performance by skewing distributions or overly emphasizing old knowledge
    \item too many synthesized features can cause the estimated distribution to skew and reduce the accuracy
\end{itemize}
\newpage
\subsection{CLIP}
Initial results of CLIP model \cite{pmlr-v139-radford21a} are impressive showing that it achieves with ViT-L/14 as backbone SOTA performance on 21 of 27 datasets tested.

On Zero-Shot transfer clip obtains best results on three different datasets, on Yahoo obtains 98.4\% compared to Visual N-Grams that scores only 72.4, then on ImageNet obtains 76.2\% compared to other one that achieves only 11.5\% and finally on SUN it achieves 58.5\%, other model obtaining 23\%.

CLIP performance varies significantly depending on the domain and type of text so:
\begin{itemize}
    \item obtains very good results on Hateful Memes, SST-2 datasets
    \item performance lowers on datasets involving handwritten and street view numbers like MNIST or SVHN
    \item performance gets very low on zero-shot MNIST
\end{itemize}
CLIP improves zero-shot retrieval and is competitive with the best fine-tuned results on Flickr30k and MSCOCO datasets. 

We figured several flaws for CLIP and will present bellow:
\begin{itemize}
    \item sensitivity to the domain and type of text
    \item struggles with low-resolution and blurry images
    \item struggles with datasets involving handwritten numbers and street view numbers
\end{itemize}
\subsection{CLOOB}
Initial results with CLOOB show that when compared to CLIP it obtains better results on 7 out of 8 datasets for zero-shot including on ImageNet and ImageNet V2.

Another result show that with both models trained on YFCC with ResNet-50 encoder, CLOOB consistently outperforms CLIP at all tasks with small exceptions that appear for linear probing dataset. When encoder size is increased for example using ResNet50x4 CLOOB outperforms CLIP on all tasks with a good margin on some of datasets. 

Another significant result also shows that CLOOB clearly outperforms CLIP in both image-to-text and text-to-image retrieval tasks on the CC validation set containing 13,330 samples.

During the study we also identified few flaws with CLOOB that will be presented bellow:
\begin{itemize}
    \item mixed results when comparing CLOOB and CLIP with different ResNet encoders
    \item textual descriptions in the YFCC dataset contain superfluous information due to the lack of filtering by quality
    \item CLOOB doesn't perform so well when using long sized prompts
\end{itemize}
\section{Main results and Comparison}
\subsection{Comparison with initial approach}
A first comparison will be made with the initial approach that uses CLIP but we change the backbone from ViT-B/16 to ResNet50 and ResNet50x4 for a more fair comparison, since we do not have pretrained versions of CLOOB with ViT-B/16. We decided to use for this comparison only three sessions to reduce the time needed for different tests.
In Tab. \ref{tab:rn50} we can see the first comparison results for ResNet50 backbone.
\begin{table}[ht]
\centering
\begin{tabular}{lcccc}
\toprule
\textbf{Metric} & \textbf{Session} & \textbf{LP-DiF + Clip} & \textbf{LP-DiF + Cloob} \\
\midrule
\textbf{Train Accuracy} & 0 & 42 & 40 \\
                     & 1 & 90 & 85 \\
                     & 2 & 91 & 87 \\
\midrule
\textbf{Train Loss} & 0 & 2.1 & 2.25 \\
                     & 1 & 0.25 & 0.4 \\
                     & 2 & 0.25 & 0.3 \\
\midrule
\textbf{Validation Accuracy} & 0 & 57 & 56 \\
                             & 1 & 55 & 52 \\
                             & 2 & 50 & 52 \\
\midrule
\textbf{Validation Error rate} & 0 & 44 & 42 \\
                               & 1 & 75 & 76 \\
                               & 2 & 47 & 80 \\
\bottomrule
\end{tabular}
\caption{Performance for RN50 backbone across different sessions}
\label{tab:rn50}
\end{table}

As we can see the initial results with the smaller encoder ResNet50, the original approach obtains better results than ours with CLOOB but the difference is rather small so we continue with a bigger encoder to see if we can get better results.
\newline
Now in Tab. \ref{tab:rn50x4} we can see the comparison results using ResNet50x4 backbone.
\newpage
\begin{table}[htbp]
\centering
\begin{tabular}{lcccc}
\toprule
\textbf{Metric} & \textbf{Session} & \textbf{LP-DiF + Clip} & \textbf{LP-DiF + Cloob} \\
\midrule
\textbf{Train Accuracy} & 0 & 32 & 40 \\
                        & 1 & 85 & 90 \\
                        & 2 & 95 & 92 \\
\midrule
\textbf{Train Loss} & 0 & 2.6 & 2.2 \\
                    & 1 & 0.3 & 0.25 \\
                    & 2 & 0.2 & 0.25 \\
\midrule
\textbf{Validation Accuracy} & 0 & 42 & 55 \\
                             & 1 & 42 & 52 \\
                             & 2 & 42 & 49 \\
\midrule
\textbf{Validation Error rate} & 0 & 57 & 43 \\
                               & 1 & 80 & 75 \\
                               & 2 & 78 & 76 \\
\bottomrule
\end{tabular}
\caption{Performance metrics for RN50x4 across different sessions}
\label{tab:rn50x4}
\end{table}
As we supposed, having in mind models architectures and experiments made with CLOOB framework, CLOOB perfomance improves when increasing encoder size, how it starts to outperform CLIP in most cases.
\subsection{Compare with SOTA}
Now to compare with other SOTA models we'll shortly present them.
\subsubsection{Semantic-visual Guided Transformer for Few-shot Class-incremental Learning}
The proposed solution \cite{qiu2023semantic} is a Semantic-Visual Guided Transformer (SV-T) designed to enhance feature extraction for FSCIL. The SV-T integrates both visual and semantic information to improve the robustness of the pre-trained feature backbone.

The core components of the framework are:
\begin{itemize}
    \item Visual Labels (supervised by visual labels provided by the base classes to optimize the Transformer)
    \item Text Encoder (automatically generates corresponding semantic labels for each image from the base classes)
    \item Guidance (uses these semantic labels to guide the hyperparameter updates of the Transformer, ensuring better generalization and reduced overfitting)
\end{itemize}
The results of the proposed SV-T approach were validated through extensive experiments on three benchmarks, two FSCIL architectures, and two Transformer variants. The results demonstrated significant improvements over state-of-the-art FSCIL methods. On CUB200 Dataset obtains 76.17\% accuracy with 12.43\% more than previous SOTA, and on mini-Imagenet shows 81.65\% accuracy with 22.08\% more than previous SOTA.

Advantages of this framework include:
\begin{itemize}
    \item Enhanced Feature Extraction (utilizing both visual and semantic labels, improves accuracy of feature extraction)
    \item Flexibility (is a simple and independent module that can be integrated into various existing FSCIL architectures and Transformer variants)
    \item Reduced Overfitting (the dual-guidance mechanism helps mitigate the risk of overfitting)
    \item Improved Generalization (combined use of semantic and visual features provides a stronger generalization capability) 
\end{itemize}
\subsubsection{Few-Shot Class-Incremental Learning by Sampling Multi-Phase Tasks}
\begin{figure}[ht]
    \centering
    \includegraphics[width=\columnwidth]{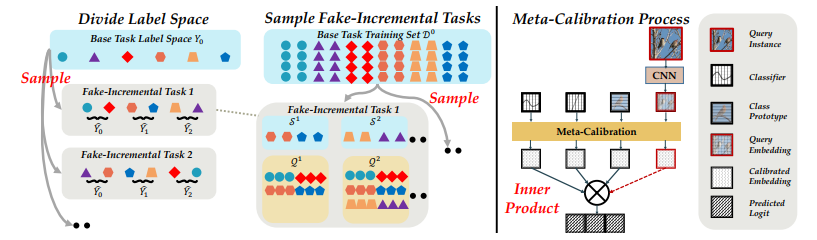}
    \caption{Illustration of LIMIT. Left: We sample fake-incremental tasks from the base training set D0
, forming various fake-tasks.
Right: Meta-calibration process. The model needs to calibrate between old class classifiers and new class prototypes with a set-to-set
function. We also input the query instance embedding into the meta-calibration module to contextualize the classification task, generating
instance-specific embeddings. The final logit is calculated by the inner-product of the transformed classifier and transformed query
embedding}
    \label{Figure2}
\end{figure}

The authors propose a solution \cite{FS_CIL_SamplingMultiPhaseTasks2022-daweizhou} (Fig. \ref{Figure2}) that works by learning multi-phase incremental tasks (LIMIT), which is a new paradigm based on meta-learning that synthesizes fake FSCIL tasks from the base dataset and prepares the model for future updates. Solution involves meta-training the model on these synthetic tasks and a calibration module based on the transformer architecture that serves to bridge the semantic gap between old class classifiers and new class prototypes. This module adapts the instance-specific embeddings to ensure they are on the same scale, enhancing the model's ability to generalize to new classes while retaining knowledge of old ones. 
The steps to obtain the results are as follows:
\begin{itemize}
    \item Pre-train the model with the base dataset using cross-entropy loss.
    \item Meta-train the model with the fake-incremental tasks using the proposed training protocol and meta-calibration module.
    \item Meta-test the model with the real incremental tasks using the same procedure as meta-training, but without back-propagation.
\end{itemize}
The model is created by combining a convolutional neural network (CNN) as the embedding function and a linear layer as the classifier.

The results were obtained by conducting experiments on three benchmark datasets (CIFAR100, miniImageNet, and CUB200) and a large-scale dataset (ImageNet ILSVRC2012). It shows that LIMIT achieves SOTA performance in FSCIL tasks, the evaluation metrics included accuracy in each incremental learning session, showing significant improvements over previous methods. The ablation study on CIFAR100 highlighted that each component of LIMIT contributed to enhanced performance, with the complete model exhibiting superior accuracy and lower forgetting rates compared to other approaches.

The main advantages of LIMIT are:
\begin{itemize}
    \item Generalizability (training on synthetic tasks allows the model to learn a generalizable feature space)
    \item Calibration Module (improve the model's ability to handle new classes without forgetting old ones)
    \item Efficiency (does not require back-propagation, making it efficient for real-time applications)
    \item Flexibility (it can handle varying numbers of new classes and instances per class without structural changes to the model)
    \item Reproducibility (provides detailed method description, pseudo code, source code, and dataset splits on GitHub)
\end{itemize}
\subsubsection{Constructing Sample-to-Class Graph for Few-Shot Class-Incremental Learning}
\begin{figure}[ht]
    \centering
    \includegraphics[width=\columnwidth]{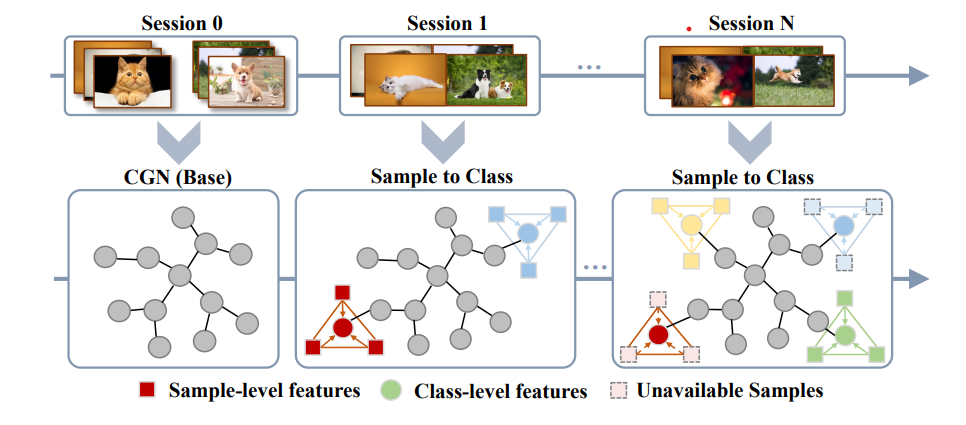}
    \caption{Illustration of our proposed S2C for FSCIL. Top: the setting of FSCIL. Bottom: Sample-level to Class-level graphs.}
    \label{Figure3}
\end{figure}

The authors propose a solution (Fig. \ref{Figure3}) \cite{hu2023constructing} that consists of a Sample-to-Class (S2C) graph learning method for FSCIL, that contains a Sample-level Graph Network (SGN) and a Class-level Graph Network (CGN).

The SGN is designed to explore relationships among samples within a single few-shot session, refining features and reducing overfitting by clustering samples from the same class and distinguishing those from different classes. The CGN extends this by establishing connections between old and new class-level features, thereby capturing contextual relationships across different sessions.

The experimental results demonstrate that the S2C model outperforms state-of-the-art methods in FSCIL tasks on benchmark datasets such as MiniImageNet (0.82\% accuracy more than SOTA), CIFAR100 (0.39\% accuracy more than SOTA), and CUB-200-2011 (0.5\% accuracy more than SOTA). The ablation studies reveal that incorporating both SGN and CGN significantly improves performance. Visualization of incremental sessions using t-SNE shows that S2C effectively adapts prototypes and refines decision boundaries, maintaining distinguishable class-level attributes even with limited samples.

The advantages of the approach are:
\begin{itemize}
    \item Enhanced Feature Extraction (refines sample features by exploring inter-sample relationships, which reduces overfitting)
    \item Contextual Relationship Modeling (captures and adjusts relationships between old and new classes, ensuring better adaptation in incremental learning)
    \item Improved Performance (obtains SOTA results on several datasets)
    \item Visualization Support (the use of t-SNE for visualizing decision boundaries highlights the effectiveness of S2C in managing both old and new classes)
    \item Reproducibility (provides hyperparameters, data augmentation and source code on GitHub)  
\end{itemize}
\subsubsection{Main comparison}
\begin{figure}[ht]
    \centering
    \adjustbox{max width=1.2\linewidth, center}{
        \includegraphics[width=1.2\textwidth]{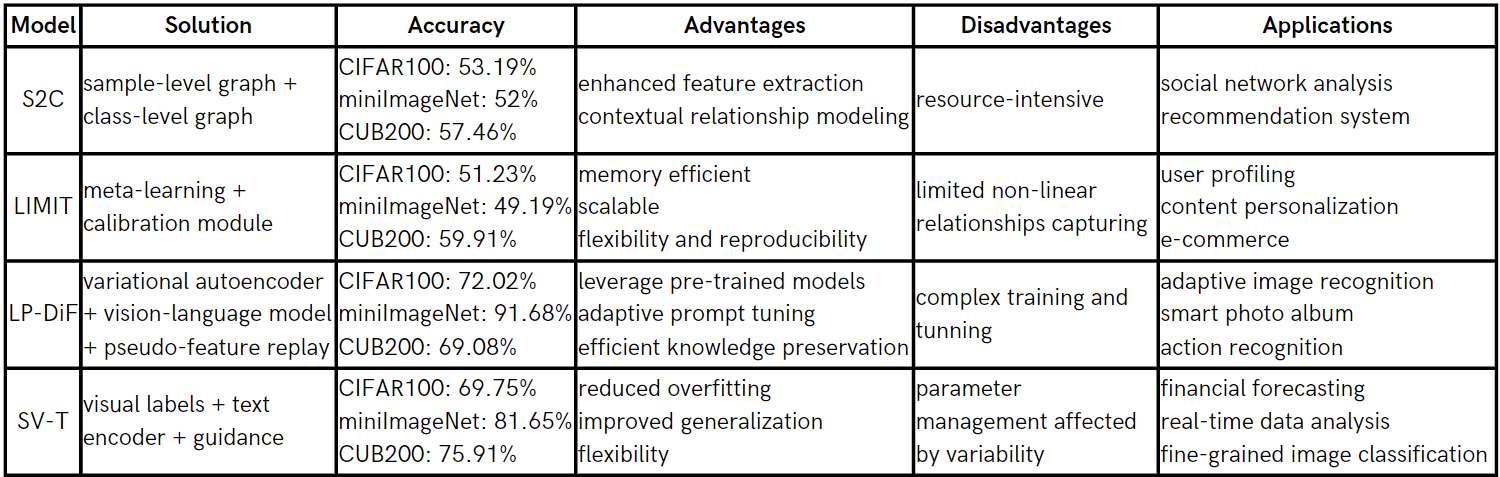}
    }
    \caption{FSCIL approaches comparison}
    \label{Figure4}
\end{figure}

In this part we will compare some of the best approaches on FSCIL (Fig. \ref{Figure4}) using accuracy as the main metric, we'll highlight some of advantages and disadvantages for each of them along some applications where their capabilities will shine.
\newpage
We can see that LP-DiF obtains best overall results on compared datasets by a good margin and with our approach it should further improve its performance.
\section{Discussion}
\subsection{Conclusion}
In this paper, we have explored the challenges and potential solutions within the realm of few-shot class-incremental learning (FSCIL). Our research delves into the core limitations of current methodologies, such as catastrophic forgetting, class imbalance, and the need for improved evaluation protocols. Through our work, we aim to mitigate these issues by incorporating advanced techniques and lay the groundwork for advancements in the field.

Our findings suggest that while significant progress has been made, there is still room for future advancements. We have identified several key areas for further investigation, including the development of new benchmark datasets (similar to those that appear in LP-DiF), enhancement of model generalization capabilities, and exploration of task-free incremental learning settings. Additionally, we propose approaches such as hybrid models and novel network architectures to address specific limitations such as the saturation effect in Hopfield networks and the performance of CLIP on fine-grained classification tasks.

In conclusion, our research highlights the importance of ongoing innovation and interdisciplinary collaboration in advancing the field of few-shot class-incremental learning. By building on the foundational work presented in this paper, we anticipate significant strides in developing more effective and efficient learning algorithms, ultimately contributing to the broader landscape of artificial intelligence.
\subsection{Limitations and open problems}
The main open problem that remains for our approach is the backbone model used, we only tried ResNet50 and ResNet50x4, when there are other proven better methods such as ViT-B in his different variants. The training dataset of CLOOB and CLIP models can also represent a path for further improvement, since the models that we used were trained on YFCC dataset when there are others that can prove to be better.
Even trough we analyzed several approaches and picked the most promising one that managed to mitigate the catastrophic forgetting and obtained very good results, we still found some weaknesses that would need some further study and innovative approaches to be fixed. Some of those are:
\begin{itemize}
    \item despite various strategies, mitigating catastrophic forgetting remains a significant challenge
    \item FSCIL methods often struggle with class imbalance
    \item new evaluation protocols and metrics that can capture the complexity and diversity
    \item there is a need for methods that provide interpretable incremental learning processes
    \item encoder size has a huge effect on performance for CLOOB
    \item Hopfield networks increase the saturation effect of the InfoNCE objective which affects learning
    \item CLIP struggles with fine-grained classification tasks
    \item for tasks not included in CLIP pre-training data performance tend to be random
\end{itemize}
\subsection{Future work directions}
During our work on this approach we also managed to put up some future work directions that could help obtaining a more reliable and complete solution to FSCIL challenges that would fix some of the open problem found but also extend the application to other fields. Some of the points that we propose are:
\begin{itemize}
    \item develop adaptive sampling strategies that dynamically adjust to class imbalances
    \item investigate the trade-offs between model complexity and performance in various FSCIL tasks
    \item extending class incremental learning to other domains and modalities, such as natural language processing, speech recognition, reinforcement learning, and multi-modal learning
    \item combining the methods for mitigating catastrophic forgetting with methods for increasing generalization capacity of the model
    \item establish protocols for systematic adaptation of pre-trained models to new and unseen tasks
    \item use different backbone architectures of different sizes and analyze performance improvement
    \item test the proposed approach when using ViT-B/16 backbone and compare with SOTA
    \item train CLOOB on different datasets and with different encoders to see how performance evolves
    \item exploring exemplar learning, feature rehearsal, self- and unsupervised incremental learning, new loss functions beyond cross-entropy, meta-learning, and task-free settings
\end{itemize}
The proposed future work directions aim to extend the applicability of FSCIL to various domains, enhance model robustness, and ensure continuous learning from dynamic data streams to replace traditional models like \cite{onchis2014observing} and \cite{feichtinger2010constructive}, based on signal processing. By addressing these challenges, we hope to pave the way for more resilient and adaptable learning systems that can effectively handle the complexities of real-world applications.
As a final remark, this paper presented an overview of the current state of the art in FACIL domain, a new approach that leveraged two innovative methods to obtain better results and improve the performance and finally provided inspiring and guiding future research direction in this new and promising area of study.
\bibliographystyle{plain}
\bibliography{biblio}
\end{document}